%% file: main.tex
\documentclass[sigconf, screen, dvipsnames, nonacm]{acmart}
\usepackage{hyperref}
\usepackage[noend]{algorithmic}
\usepackage{algorithm}
\usepackage{multirow}
\usepackage{amsmath}
\usepackage{bbm}
\usepackage{booktabs}
\usepackage[inline]{enumitem} % Inline lists
\usepackage{subcaption}
\usepackage{bm} % Bold greek
\usepackage{xcolor}
\usepackage{balance}
\usepackage{lipsum}

\usepackage{wrapfig}

\usepackage{arydshln}
\makeatletter
\def\adl@drawiv#1#2#3{
        \hskip.5\tabcolsep
        \xleaders#3{#2.5\@tempdimb #1{1}#2.5\@tempdimb}%
                #2\z@ plus1fil minus1fil\relax
        \hskip.5\tabcolsep}
\newcommand{\cdashlinelr}[1]{%
  \noalign{\vskip\aboverulesep
          \global\let\@dashdrawstore\adl@draw
          \global\let\ adl@draw\adl@drawiv}
  \cdashline{#1}
  \noalign{\global\let\adl@draw\@dashdrawstore
          \vskip\belowrulesep}}
\makeatother

\usepackage{tikz}
\usetikzlibrary{automata, arrows, bayesnet, bending}

\usepackage{tikzscale}

\begin{document}
\title{Unifying On- and Off-Policy Variance Reduction Methods}

\author{Olivier Jeunen}
\affiliation{
  \institution{aampe}
  \city{Antwerp}
  \country{Belgium}
}
\email{olivier@aampe.com}

\begin{abstract}
Continuous and efficient experimentation is key to the practical success of user-facing applications on the web, both through online A/B-tests and off-policy evaluation.
Despite their shared objective---estimating the incremental value of a treatment---these domains often operate in isolation, utilising distinct terminologies and statistical toolkits.
This paper bridges that divide by establishing a formal equivalence between their canonical variance reduction methods.

We prove that the standard online Difference-in-Means estimator is mathematically identical to an off-policy Inverse Propensity Scoring  estimator equipped with an optimal (variance-minimising) additive control variate.
Extending this unification, we demonstrate that widespread regression adjustment methods (such as CUPED, CUPAC, and ML-RATE) are structurally equivalent to Doubly Robust estimation.
This unified view extends our understanding of commonly used approaches, and can guide practitioners and researchers working on either class of problems. 
\end{abstract}

\maketitle

\input{1.Introduction}
\input{2.Methodology}
\input{3.Conclusions}

\balance

\bibliographystyle{ACM-Reference-Format}
\bibliography{bibliography}

\end{document}

%% file: 1.Introduction.tex
\section{Introduction \& Motivation}
Personalisation powers customer engagement on the web, fuelled by continuous experimentation that empowers researchers and practitioners to make data-driven decisions.
The landscape of practical experimentation methods is currently bifurcated into two distinct paradigms: online experimentation through A/B-testing~\cite{kohavi2020trustworthy}, and offline experiments through Off-Policy Evaluation (OPE)~\cite{Saito2022}.

In the online world, practitioners rely on randomised assignments to estimate treatment effects.
The standard toolkit includes the Difference-in-Means estimator, with variance reduction techniques that leverage additive control variates in the form of regression adjustments (e.g. CUPED~\cite{Deng2013}, CUPAC~\cite{Li2020_cupac}, and ML-RATE~\cite{Guo2021}).

The high engineering, opportunity and business costs of running online experiments motivate the need for offline alternatives~\cite{Gilotte2018}.
These rely on counterfactual inference using logged data~\cite{Vasile2020,Bottou2013}.
The Inverse Propensity Scoring (IPS) or Horvitz--Thompson~\cite{HorvitzThompson1952} estimator is foundational for this class of experiments.
Here too, additive control variates are often used to reduce variance and, as a result, improve statistical power~\cite{Dudik2014,Farajtabar2018,Gupta2024,Jeunen2024_DeltaOPE,jeunen2026additivecontrolvariatesdominate}.

Whilst these two domains share the same fundamental goal---estimating the causal impact of a policy deployment with minimal variance---they often operate with disparate terminologies, seemingly disjoint methods, and separated engineering stacks.
An overabundance of methodological decisions then often obfuscates the practical work to be done, leading to fragmented infrastructure and preventing the cross-pollination of advancements in variance reduction techniques for either class of problems.

This work presents a unified view of on- and off-policy experimentation.
We demonstrate that the divide between these modes is largely artificial.
Specifically, we derive two key equivalences:
\begin{enumerate}
    \item \textbf{DiM $\equiv$ $\beta^{\star}$IPS}. We show that the standard online Difference-in-Means (DiM) estimator is mathematically equivalent to an off-policy IPS estimator augmented with a recently proposed optimal additive control variate $\beta^\star$~\cite{Gupta2024, Jeunen2024_DeltaOPE}.
    \item \textbf{CUPED, CUPAC, ML-RATE $\equiv$ Doubly Robust}. We demonstrate that regression-adjusted online estimators~\cite{Deng2013, Baweja2024, Li2020_cupac,Guo2021} are structurally equivalent to Doubly Robust estimators~\cite{Dudik2014,Farajtabar2018}, where the reward model is action-agnostic.
\end{enumerate}

Importance sampling~\cite[Ch. 9]{Owen2013}, regression adjustment~\cite{Freedman2008}, and doubly robust estimation~\cite{Dudik2014} are all well-studied in isolation.
Whilst the correspondence between regression adjustment (ANCOVA) and weighted estimators is established in the broader causal inference literature~\cite{imbens2015causal}, this connection has not been operationalised for the distinct terminologies, finite-sample variance estimators, and implementation practices prevalent in online and offline evaluation of Information Retrieval and RecSys applications, which have historically developed independently.
Our contribution is to close this gap: we formally prove that commonly used estimators in these two communities are exactly the same objects under different parameterisations, establishing equality not only in expectation but with finite samples.
Concretely, this unification enables direct cross-pollination of techniques across these communities, and surfaces a non-obvious degrees-of-freedom correction with immediate consequences for practical variance estimation.

From a theoretical perspective, this unified view on existing approaches highlights underlying complementarities and overlap whilst exposing gaps that could inform fruitful areas for future research.
Beyond the value of theoretical insights, we discuss the practical utility and impact of our findings on implementation details, such as degrees-of-freedom corrections.

%% file: 2.Methodology.tex
\section{Methodology, Background \& Notation}
We conceptualise personalised treatment regimes as \emph{policies}~\cite{Vasile2020}.
A policy defines a probability distribution over actions $A$, conditioned on a context $X$.
Contextual covariates can be multitude: a user identifier, device type, accumulating in-session data, et cetera.
Actions are general as well: a ranking~\cite{Gupta2024_WSDM}, an item recommendation~\cite{Joachims2021}, a sequence of tokens~\cite{Ouyang2022}, et cetera.\footnote{This general framing subsumes simple A/B-testing (i.e. no covariates and a binary treatment), and in no way limits the broad applicability of our exposition below.}
A policy maps a context $x$ to a probability distribution over $A$ as $\pi(a|x) \equiv \mathsf{P}(A=a|X=x;\Pi=\pi)$.

We measure \emph{rewards} or \emph{outcomes} occurring under varying policies: these correspond to metrics used in online experimentation (e.g. session time, clicks, revenue)~\cite{Jeunen2024_RecSysIndustry}.
The \emph{value} of a policy is the expectation of the outcomes we obtain by deploying it:
\[
V(\pi) = \mathop{\mathbb{E}}\limits_{x \sim \mathsf{P}(X)}\mathop{\mathbb{E}}\limits_{a \sim \pi(\cdot|x)}\mathop{\mathbb{E}}\limits_{\mathsf{P}(Y|X=x;A=a)}\left[Y\right].
\]

The real estimand of interest is typically an Average Treatment Effect (ATE) on policy values~\cite{kohavi2020trustworthy,Jeunen2024_DeltaOPE}.
For a pair of policies $\pi, \pi^{\prime}$:
\[
V_{\Delta}(\pi,\pi^{\prime}) = V(\pi) - V(\pi^{\prime}).
\]

\subsection{Online Controlled Experiments}
In an online experiment, we deploy both policies to the user population and directly target the ATE by separating the sample:
\[
V_{\Delta}(\pi,\pi^{\prime}) = \underbrace{\mathop{\mathbb{E}}\limits_{a\sim\pi(\cdot|x)}[Y]}_{V(\pi)} - \underbrace{\mathop{\mathbb{E}}\limits_{a\sim\pi^{\prime}(\cdot|x)}[Y]}_{V(\pi^{\prime})}.
\]

For a dataset $\mathcal{D} =\{(x_i, a_i, y_i, \pi_i)\}_{i=1}^{N}$,  we denote a subset generated under policy $\pi$ as  $\mathcal{D}_\pi =\{(x_i, a_i, y_i, \pi_i) \in \mathcal{D}|\pi_i=\pi\}$.

\paragraph{The Difference-in-Means (DiM) estimator}
Straightforwardly, a difference in sample means give rise to an unbiased ATE estimate:
\begin{gather}
\hat\mu(Y,\pi) = \frac{1}{|\mathcal{D_{\pi}}|} \sum_{(x_i,a_i,y_i) \in \mathcal{D}_{\pi}} y_i,\\
\hat V_{\Delta-{\rm DiM}}(\pi, \pi^{\prime}) =\hat \mu(Y,\pi)-\hat \mu(Y,\pi^{\prime}).
\end{gather}
Its variance is straightforwardly computed by combining the two independent samples.
With the sample variances computed as:
\[
\hat\sigma^2(Y,\pi) = \frac{1}{|\mathcal{D}_{\pi}| - 1}\sum_{(x_i,a_i,y_i)\in \mathcal{D}_\pi} \left( y_i - \hat\mu(Y,\pi) \right)^2.
\]
The variance of the mean of the DiM estimator is:
\begin{equation}    
\hat{\mathrm{Var}}\big[\hat V_{\Delta\text{-DiM}}(\pi,\pi^{\prime})\big]= \frac{\hat\sigma^{2}(Y,{\pi})}{|\mathcal{D}_{\pi}|}   + \frac{ \hat\sigma^{2}(Y,{\pi^{\prime}})}{|\mathcal{D}_{\pi^{\prime}}|}.
\end{equation}

This standard statistical machinery is the go-to approach to estimate A/B-test outcomes, providing an unbiased estimate of the treatment effect.
Nevertheless, high variance in the outcome $Y$ is likely to propagate and hamper statistical power.
In many cases, this variance in the outcome can partially be explained by covariates and not by the treatment.
When that applies, we can reduce variance in the outcome estimate by removing variations caused by covariates~\cite{Freedman2008}. 

\paragraph{Regression Adjustments}
Assume we have access to a model that maps covariates to outcomes: $f(X) \approx Y$.
Since $f(X)$ depends only on the context $X$ and not the action $A$, its expectation is invariant to the tested policies. 
Then, we can obtain a regression-adjusted unbiased decomposition of our estimand as:
\begin{align}
     \mathop{\mathbb{E}}\limits_{\substack{x\sim \mathsf{P}(X)\\a\sim\pi(\cdot|x)}}[Y - f(X)] \quad-&\quad   \mathop{\mathbb{E}}\limits_{\substack{x\sim \mathsf{P}(X)\\a\sim\pi'(\cdot|x)}}[Y - f(X)] \nonumber \\
= \left(\mathop{\mathbb{E}}\limits_{\substack{x\sim \mathsf{P}(X)\\a\sim\pi(\cdot|x)}}[Y] - \mathop{\mathbb{E}}\limits_{x\sim \mathsf{P}(X)}[f(X)]\right) -& \left(\mathop{\mathbb{E}}\limits_{\substack{x\sim \mathsf{P}(X)\\a\sim\pi'(\cdot|x)}}[Y] -  \mathop{\mathbb{E}}\limits_{x\sim \mathsf{P}(X)}[f(X)]\right) \nonumber \\
= \mathop{\mathbb{E}}\limits_{\substack{x\sim \mathsf{P}(X)\\a\sim\pi(\cdot|x)}}[Y] -& \mathop{\mathbb{E}}\limits_{\substack{x\sim \mathsf{P}(X)\\a\sim\pi'(\cdot|x)}}[Y] =   V_{\Delta}(\pi,\pi^{\prime}) . \nonumber
\end{align}

This retains unbiasedness with the potential of significantly reducing estimation variance, if the predictions $f(X)$ have a high correlation with the observed outcomes $Y$~\cite{Freedman2008}.
This general notation subsumes several commonly used methods.
When $f(X)$ represents pre-experiment values of $Y$, we recover CUPED~\cite{Deng2013}.
With general machine learning models for $f$, we recover CUPAC~\cite{Li2020_cupac} and ML-RATE~\cite{Guo2021}.
Additionally, any discrepancy in the average of the $f(X)$ terms between the two A/B-groups is purely driven by random sampling noise---often referred to as \emph{pre-experiment bias}---which is then effectively removed.
A regression-adjusted sample mean and variance are given by:
\begin{gather}
\hat\mu_f(Y,\pi) = \frac{1}{|\mathcal{D_{\pi}}|} \sum_{(x_i,a_i,y_i) \in \mathcal{D}_{\pi}} \left( y_i - f(x_i) \right), \\
\hat\sigma_f^2(Y,\pi) = \frac{1}{|\mathcal{D}_{\pi}| - 1}\sum_{(x_i,a_i,y_i)\in \mathcal{D}_\pi} \left( (y_i - f(x_i))- \hat\mu_f(Y,\pi) \right)^2.    
\end{gather}
With a regression-adjusted DiM estimate for the ATE as:
\begin{equation}
    \hat V_{\Delta-{\rm RADiM}}(\pi, \pi^{\prime}) =\hat \mu_f(Y,\pi)-\hat \mu_f(Y,\pi^{\prime}).
\end{equation}
The variance of the mean of the RADiM estimator is:
\begin{gather}
\hat{\mathrm{Var}}\big[\hat V_{\Delta\text{-RADiM}}(\pi,\pi^{\prime})\big]=  \frac{\hat\sigma^{2}_f({Y,\pi})}{|\mathcal{D}_{\pi}|}  + \frac{\hat\sigma^{2}_f(Y,{\pi^{\prime}})}{|\mathcal{D}_{\pi^{\prime}}|} .
\end{gather}

\paragraph{Minimising Estimation Variance}
Focusing on a single policy to reduce notational clutter, the variance of the adjusted outcome is:
\[
    {\rm Var}(Y-f(X)) = {\rm Var}(Y)+{\rm Var}(f(X))-2{\rm Cov}(Y,f(X)).
\]
Assuming $f(X)$ is scaled well, this can be written as:
\begin{align*}{\rm Var}(Y-f(X))  &= {\rm Var}(Y) - \frac{{\rm Cov}(Y,f(X))^2}{{\rm Var}(f(X)) } \\ &= {\rm Var}(Y)(1-\rho^2(Y,f(X))). \end{align*}

Here, $\rho^{2}$ represents the correlation between the outcomes and the model.
As such: the regression model $f(X)$ that minimises variance, is the one that maximises the correlation with $Y$.
This implies the model is allowed to be biased without affecting variance.
Because the regression estimates appear on both sides of the RADiM estimate, any bias also cancels out in expectation. 

% \paragraph{A Note on Difference-in-Differences}
% Interestingly, this mathematical form resembles that of a Difference-in-Differences (DiD) estimator~\cite{Athey2006}, though their motivations diverge.
% Structurally, both methods employ a nested subtraction.
% In a DiD estimator, the ``\emph{inner difference}'' typically subtracts a pre-intervention outcome from a post-intervention outcome to control for temporal trends.
% In RADiM, we subtract a predicted counterfactual $f(X)$ from an observed outcome $Y$ to control for sampling noise.
% The distinction lies in the ``\emph{outer difference}'' (Treatment vs. Control).
% DiD relies on an imperfect control group and invokes a \emph{parallel trends assumption} to identify a causal effect from observational data (removing bias whilst potentially reducing variance).
% In contrast, RADiM leverages a valid randomised control group where the effect is already identified; the adjustment serves purely to minimise the standard error (variance reduction).

\subsection{Offline Experimentation and OPE}
Deploying a policy is costly.
This has been well-documented in the literature~\cite{Gilotte2018}.
Common reasons to avoid deployment as a pre-requisite to experimental validation are:
\begin{enumerate*}[label=(\roman*)]
    \item high engineering cost to bring a prototype to production standard,
    \item high opportunity cost because of limited bandwidth and experiment duration, 
    \item business cost by exposing end users to a suboptimal policy, and
    \item lack of access to reliable large-scale experimentation infrastructure for many researchers (especially in academia).
\end{enumerate*}

As such, we often wish to leverage a large pre-collected dataset to answer questions about (differences in) \emph{counterfactual} policy values in an entirely offline manner~\cite{Bottou2013}.
The typical approach here involves the use of inverse probability weights to adjust for the difference between the sampling and target distributions.

\paragraph{IPS-weighted Estimation}
The sampling distribution is often defined by the \emph{logging} policy $\pi_0$, and we want to make inferences about the target distribution defined by policy $\pi$. 

A common approach is to use the Inverse Propensity Scoring (IPS) weights to emulate sampling from the target distributions directly.
For the ATE, the estimand under $\pi_0$ becomes:
\[
V_{\Delta}(\pi,\pi^{\prime}) = \mathop{\mathbb{E}}\limits_{a \sim \pi_0(\cdot|x)}\left[\frac{\pi(a|x) - \pi^\prime(a|x)}{\pi_0(a|x)} Y\right].
\]  
This gives rise to the standard $\Delta$-IPS estimator~\cite{Jeunen2024_DeltaOPE}:
\begin{equation}
\hat V_{\Delta{\rm-IPS}}(\pi,\pi^{\prime}) = \frac{1}{|\mathcal{D}|}\sum_{(x,a,y) \in \mathcal{D}} \frac{\pi(a|x)- \pi^\prime(a|x)}{\pi_0(a|x)} y,
\end{equation}
with sample variance:
\begin{multline}
\hat{\mathrm{Var}}\big[\hat V_{\Delta\text{-IPS}}(\pi,\pi^{\prime})\big]= \frac{1}{|\mathcal{D}|\cdot(|\mathcal{D}| - 1)} \\
\sum_{(x_i,a_i,y_i)\in \mathcal{D}} \left( \frac{\pi(a_i|x_i) - \pi^\prime(a_i|x_i)}{\pi_0(a_i|x_i)} y_i - \hat V_{\Delta\text{-IPS}}(\pi,\pi^{\prime}) \right)^2. 
\end{multline}

In practical settings, $V_{\rm \Delta-{\rm IPS}}$ can have high estimation variance. Several methods to reduce the estimation variance of $V_{\rm \Delta-{\rm IPS}}$ whilst remaining unbiased, have been proposed in the research literature.
Among those, the use of an additive control variate $\beta$ has been proven effective.
This gives rise to the $\Delta\beta$-IPS estimator:
$$
\hat V_{\Delta\beta{\rm-IPS}}(\pi,\pi^{\prime}) = \frac{1}{|\mathcal{D}|}\sum_{(x,a,y) \in \mathcal{D}} \frac{\pi(a|x)- \pi^\prime(a|x)}{\pi_0(a|x)} (y-\beta).
$$
The optimal (variance-minimising) $\beta$ is then~\cite[Eq.~21]{Jeunen2024_DeltaOPE}:
$$
\beta^{\star} = \frac{\mathbb{E}\left[\left(\frac{\pi(a|x)- \pi^\prime(a|x)}{\pi_0(a|x)}\right)^2y\right]}{\mathbb{E}\left[\left(\frac{\pi(a|x)- \pi^\prime(a|x)}{\pi_0(a|x)}\right)^2\right]}.
$$
This quantity is typically estimated from data. 
The $\hat V_{\Delta \beta^{\star}{\rm -IPS}}$ estimator can be seen as an off-policy counterpart to the online $\hat V_{\Delta{\rm-DiM}}$ estimator, as we will solidify in the next Section.

\section{Unifying Modes of Experimentation}
%%%%%%%%%%%%%%%%%%%%%%%%%%%%%%%%%%%%%%%%%%%%%

We now demonstrate that the canonical online A/B-test with a Difference-in-Means (DiM) estimator can be formalised as an instantiation of importance sampling.
Furthermore, we show an exact equivalence between the $\Delta\beta^{\star}$-IPS estimator~\cite{Gupta2024,Jeunen2024_DeltaOPE} and the standard A/B-testing approach~\cite{kohavi2020trustworthy}. 

Consider an experiment where a randomisation unit $x$ is assigned to one of two policies, $\pi$ or $\pi^{\prime}$, via a logging policy $\pi_0$.
Here, we overload the notion of ``action'': rather than a content-level action (e.g. item or ranking), the action corresponds to the experimental treatment.
In a general A/B-test, $\pi_0$ assigns treatment with probability $p$: $\pi_{0}(A=\pi) = p$ and $\pi_{0}(A=\pi^{\prime}) = 1-p$.

The \emph{action} $A$ in the OPE sense corresponds to the choice of \emph{policy} to deploy, i.e. $A \in \{\pi, \pi^{\prime}\}$.
The target estimand is the difference in values between two deterministic policies: one that always chooses $\pi$ (probability 1) and one that always chooses $\pi^{\prime}$.\footnote{This framing was also adopted by \citet{Farias2023}, in the context of online experiments with SUTVA violations~\cite{Jeunen2023_Forum}.}
This implies the importance weights take the form $w(a) = \frac{\mathbb{I}(a=\pi)}{p} - \frac{\mathbb{I}(a=\pi')}{1-p}$.

If we apply the variance-minimising estimator $\hat V_{\Delta \beta^{\star}{\rm -IPS}}$ to this problem, we first determine the optimal baseline $\beta^{\star}$.
Minimising the variance with respect to $\beta$ yields the inverse-propensity weighted mean:
\[
\hat\beta^{\star} = \frac{\mathbb{E}[w^2 Y]}{\mathbb{E}[w^2]} = \frac{\frac{p\hat\mu_{\pi}}{p^2} + \frac{(1-p)\hat\mu_{\pi'}}{(1-p)^2}}{\frac{p}{p^2} + \frac{1-p}{(1-p)^2}} = (1-p)\hat\mu_{\pi} + p\hat\mu_{\pi'}.
\]
Note that for general $p$, this optimal baseline is a weighted average, not necessarily the global mean $\hat\mu(Y)$.

We now analyse the variance of this estimator.
Let the re-weighted, baseline-adjusted outcome for a single sample be $Z = w(Y - \beta^{\star})$.
Since the estimator is unbiased ($\mathbb{E}[Z]=0$), its variance is fully determined by its second moment $\mathbb{E}[Z^2]$.
Conditioning on the treatment assignment $A$, this analytically reduces to:
\[
    \mathbb{E}[Z^2] = p \cdot \left(\frac{1}{p}\right)^2 \sigma^2_{\pi} + (1-p) \cdot \left(\frac{-1}{1-p}\right)^2 \sigma^2_{\pi^{\prime}} = \frac{\sigma^2_{\pi}}{p} + \frac{\sigma^2_{\pi^{\prime}}}{1-p}.
\]
Finally, the variance of the sample mean estimator is $\frac{1}{|\mathcal{D}|}\mathbb{E}[Z^2]$, which perfectly recovers the standard DiM variance for any treatment allocation ratio $p \in [0,1]$:
\[
    \hat{\mathrm{Var}}\big[\hat V_{\Delta\beta^{\star}\text{-IPS}}\big] = \frac{\hat\sigma^2_{\pi}}{|\mathcal{D}|p} + \frac{\hat\sigma^2_{\pi^{\prime}}}{|\mathcal{D}|(1-p)} = \frac{\hat\sigma^2_{\pi}}{|\mathcal{D}_{\pi}|} + \frac{\hat\sigma^2_{\pi^{\prime}}}{|\mathcal{D}_{\pi^{\prime}}|}.
\]
Thus, the difference-in-means approach is mathematically equivalent to off-policy estimation with an optimally chosen $\beta^{\star}$.
This implies that the distinction between ``online'' and ``offline'' experimentation is largely artificial; they are simply different parameterisations of the same underlying variance structure.

\paragraph{A Note on Bessel’s Correction}
A subtle implementation detail arises when computing sample variances and confidence intervals in practice.
The go-to approach is to compute the variance of the DiM estimator by summing the variances of the two treatment arms.
Each arm's variance calculation applies Bessel's correction independently (dividing by $|\mathcal{D}_{\pi}|-1$ and $|\mathcal{D}_{\pi^{\prime}}|-1$, respectively), effectively accounting for the estimation of two sample means.
This results in a total loss of two degrees of freedom.

When implementing the $\hat V_{\Delta\beta^{\star}\text{-IPS}}$ estimator, however, one might intuitively treat it as a standard mean estimation problem for the single transformed variable $Z = w(Y - \beta^{\star})$.
Intuition would default to dividing by $|\mathcal{D}|-1$.
This yields a variance estimate that differs from the DiM result by a factor of exactly $\frac{|\mathcal{D}|-1}{|\mathcal{D}|-2}$.
The discrepancy stems from the fact that the control variate $\beta^{\star}$ is itself estimated from the data (depending on two sample means).
This estimation consumes an additional degree of freedom.
Consequently, the correct unbiased variance estimator for $\hat V_{\Delta\beta^{\star}\text{-IPS}}$ must divide by $|\mathcal{D}|-2$.
Applying this correction recovers an exact numerical match between the on-policy (DiM) and off-policy (IPS) variance estimates.

\paragraph{Off-Policy Regression Adjustments (Doubly Robust)}
Thus far, we have established that the on-policy Difference-in-Means estimator is equivalent to the off-policy $\beta$-IPS estimator for estimating an ATE, both in terms of expectations and variances.
A natural follow-up question arises: does a similar equivalence exist for the on-policy Regression-Adjusted Difference-in-Means (RADiM) estimator in the off-policy world?

The Doubly Robust (DR) estimator provides a regression-adjusted estimate for a single policy value $V(\pi)$ by combining importance sampling with a reward model $f(x,a)$~\cite{Dudik2014}:
\begin{multline}
\hat V_{\rm DR}(\pi) = \frac{1}{|\mathcal{D}|} \\\sum_{(x_i,a_i,y_i) \in \mathcal{D}} \left(\frac{\pi(a_i|x_i)}{\pi_0(a_i|x_i)}(y_i - f(x_i,a_i)) +\sum_{a_j \in \mathcal{A}} \pi(a_j|x_i)f(x_i,a_j) \right).
\end{multline}
Whilst a direct difference-estimator $\hat V_{\Delta{\rm-DR}}$ has not been proposed in the literature, it can be straightforwardly formulated as the difference of two DR estimators~\cite{Jeunen2024_DeltaOPE}:
\begin{multline*}
\hat V_{\Delta{\rm-DR}}(\pi,\pi^{\prime}) = \frac{1}{|\mathcal{D}|}\\
\sum_{(x_i,a_i,y_i) \in \mathcal{D}} \left[ \frac{\pi(a_i|x_i)-\pi^{\prime}(a_i|x_i)}{\pi_0(a_i|x_i)}(y_i - f(x_i,a_i)) \right. \\
\left. + \sum_{a_j \in \mathcal{A}} (\pi(a_j|x_i)-\pi^{\prime}(a_j|x_i))f(x_i,a_j) \right].
\end{multline*}
The structure of the reward model $f$ is important here.
In classical Doubly Robust estimation for general policies, $f(x,a)$ models the expected reward for a specific action.
This is intuitive for e.g. item recommendation use-cases where $f(x,a)$ models a click probability---but less so in the general case where $a$ refers to an arbitrary policy being tested in an online experiment. 
If we constrain the model to be \emph{action-agnostic}---i.e., $f(x,a) \equiv f(x)$, as is standard in online CUPED-style adjustments---the second term in the outer sum vanishes.
Since $\sum_{a} \pi(a|x) = \sum_{a} \pi'(a|x) = 1$, the term $\sum_{a} (\pi(a|x)-\pi'(a|x))f(x)$ simplifies to $f(x)(1-1) = 0$.

We can verify the unbiasedness of the $\hat V_{\Delta{\rm-DR}}(\pi,\pi^\prime)$ estimator:
\begin{align*}
&\mathbb{E}\left[\hat V_{\Delta-{\rm DR}}(\pi,\pi^\prime)\right] = \mathbb{E}_{x, a \sim \pi_0, y}\left[ \frac{\pi(a|x) - \pi'(a|x)}{\pi_0(a|x)} (y - f(x)) \right] \\
&= \mathbb{E}_{x}\left[ \sum_{a} \pi_0(a|x) \frac{\pi(a|x) - \pi'(a|x)}{\pi_0(a|x)} \mathbb{E}[y - f(x) \mid x, a] \right] \\
&= \mathbb{E}_{x}\left[ \sum_{a} (\pi(a|x) - \pi'(a|x)) (\mathbb{E}[Y|x,a] - f(x)) \right] \\
&= \mathbb{E}_{x}\left[ \sum_{a} (\pi(a|x)- \pi'(a|x)) \mathbb{E}[Y|x,a] - f(x)\underbrace{\sum_{a}(\pi(a|x) - \pi'(a|x))}_{0} \right] \\
&= \mathbb{E}_{x}\left[ V(\pi|x) - V(\pi'|x) \right] = V_{\Delta}(\pi,\pi^{\prime}).
\end{align*}
%%%%%%%%%%%%%%%%%%%%%%%%%%%%%%%%%%%%%%%%%%%%%%%%%%%%%%%%%%%%%
This unbiasedness result connects the mean of $\Delta$-DR to the mean of RADiM.
Next, we match their variance.
We assume the reward model $f(x)$ is centred at the variance-minimising baseline derived in the previous section (i.e. $\mathbb{E}[f(x)] = \beta^{\star} = (1-p)\mu_{\pi} + p\mu_{\pi'}$).
Note that this is trivially achievable by translating $f(x)$.
We define the weighted, adjusted outcome for a single sample as $Z = w(y - f(x))$.
Since the estimator is unbiased, $\mathrm{Var}(\hat V_{\Delta-{\rm DR}}) = \frac{1}{|\mathcal{D}|}\mathrm{Var}(Z)$.

Conditioning on the treatment assignment $A$, we compute the expected squared residual analytically.
With importance weights $w = \frac{1}{p}$ for $a=\pi$ and $w = \frac{-1}{1-p}$ for $a=\pi'$:
\begin{align}
    \mathbb{E}[Z^2] &= p  \mathbb{E}\left[ \left(\frac{y-f(x)}{p}\right)^2 \Bigg| \pi \right] + (1-p)  \mathbb{E}\left[ \left(\frac{y-f(x)}{-(1-p)}\right)^2 \Bigg| \pi' \right] \nonumber\\
    &= \frac{1}{p} \mathbb{E}[(y-f(x))^2 | \pi] + \frac{1}{1-p} \mathbb{E}[(y-f(x))^2 | \pi']. \nonumber
\end{align}
Because $f(x)$ is independent of $A$ and centred at the optimal baseline $\beta^{\star}$, the cross-terms vanish and the residual variance simplifies to the intra-group variance:
\begin{equation}
    \mathbb{E}[Z^2] = \frac{\sigma^2_{f,\pi}}{p} + \frac{\sigma^2_{f,\pi'}}{1-p}. \nonumber
\end{equation}
Finally, the variance of the sample mean estimator is:
\begin{align}
    \mathrm{Var}(\hat V_{\Delta-{\rm DR}}) &= \frac{1}{|\mathcal{D}|} \left( \frac{\sigma^2_{f,\pi}}{p} + \frac{\sigma^2_{f,\pi'}}{1-p} \right) \\
    &= \frac{\sigma^2_{f,\pi}}{|\mathcal{D}|p} + \frac{\sigma^2_{f,\pi'}}{|\mathcal{D}|(1-p)} = \frac{\sigma^2_{f,\pi}}{|\mathcal{D}_\pi|} + \frac{\sigma^2_{f,\pi'}}{|\mathcal{D}_{\pi'}|}.
\end{align}
This confirms that a Doubly Robust estimator for the ATE, when constrained to an action-agnostic reward model centred at $\beta^{\star}$, is mathematically equivalent to a Regression-Adjusted Difference-in-Means estimator (RADiM) under any treatment allocation ratio.

Note that restricting the reward model $f$ to be action-agnostic does not limit the generality of our equivalence for online experiments: such models are standard in regression-adjusted A/B-testing, while action-dependent reward models are common in off-policy evaluation but largely absent from the online experimentation literature.
This distinction highlights a natural direction for future work, namely extending online variance reduction methods to exploit action-aware reward models.
This holds promise for further variance reduction in recommendation and ranking applications.

%% file: 3.Conclusions.tex
\section{Conclusions \& Outlook}
In this work, we have bridged the gap between two disconnected experimentation paradigms: online A/B-testing and offline Off-Policy Evaluation (OPE).
By deriving formal equivalences, we demonstrated that the standard Difference-in-Means (DiM) estimator is mathematically identical to an Inverse Propensity Scoring (IPS) estimator with an optimal baseline~\cite{Gupta2024}.
Furthermore, we showed that modern regression-adjusted estimators (CUPED~\cite{Deng2013}, CUPAC~\cite{Li2020_cupac}, ML-RATE~\cite{Guo2021}) are structurally equivalent to Doubly Robust (DR) estimation where the reward model is action-agnostic~\cite{Dudik2014}.

These findings suggest that the distinction between ``online'' and ``offline'' variance reduction is largely artificial.
Practically, this unification allows for the cross-pollination of techniques: insights regarding degrees-of-freedom corrections in OPE directly apply to online settings, while advancements in online control variates can inform offline baseline construction.
Future work should explore relaxing the action-agnostic constraint in online settings, potentially leveraging full OPE estimators to exploit policy overlap for further variance reduction in both on- and offline experimental settings.